\documentclass[fontsize=10pt]{IEEEtran}
\usepackage[utf8]{inputenc}

\usepackage{mathtools}  
\usepackage{amsmath}
\usepackage{amsfonts}

\usepackage{graphicx}  
\usepackage{afterpage}
\ifCLASSOPTIONcompsoc
    \usepackage[caption=false, font=normalsize, labelfont=sf, textfont=sf]{subfig}
\else
\usepackage[caption=false, font=footnotesize]{subfig}
\fi

\usepackage{bm}  
\usepackage{soul}
\usepackage{xcolor}

\usepackage{import}  
\usepackage{cleveref}  

\usepackage[section]{algorithm} 
\usepackage{algpseudocode}

\usepackage[style=ieee]{biblatex}
\newcommand{\bbR}{\mathbb{R}}  
\DeclareMathOperator{\Tr}{Tr}  
\DeclareMathOperator{\Vectorize}{\text{vec}}

\DeclareMathOperator*{\argmin}{arg\,min}

\newcommand{\HH}{{\bm{\mathcal{H}}}}

\renewcommand{\top}{\mathsf{T}}

\newcommand{\tensor}[1]{\bm{\mathcal{#1}}}
\newcommand{\mat}[1]{\mathbf{#1}}
\newcommand{\vect}[1]{\mathbf{#1}}

\usepackage{scalerel}
\newcommand\widetildemat[1]{\stackrel{\hstretch{1.2}{\sim}}{\smash{\mathcal{#1}}\rule{0pt}{1.2ex}}}
\newcommand\widetildevec[1]{\stackrel{\hstretch{1.2}{\sim}}{\smash{\mathcal{#1}}\rule{0pt}{.9ex}}}

\newcommand{\mX}{\mat{X}}
\newcommand{\mH}{\mat{H}}
\newcommand{\mS}{\mat{S}}

\newcommand{\tW}{\tensor{W}}
\newcommand{\mW}{\mat{W}}   
\newcommand{\mWl}{\mat{W}_\ell}
\newcommand{\mWk}{\mat{W}_{::k}}

\newcommand{\vw}{\vect{w}}
\newcommand{\vh}{\vect{h}}

\begin{document}

\title{Fast Convolutive Nonnegative Matrix Factorization Through Coordinate and Block Coordinate Updates}
\author{
Anthony Degleris,
Benjamin Antin,
Surya Ganguli,
Alex H Williams%
\thanks{This work received support from the Department of Energy Computational Science Graduate Fellowship (CSGF) program, the Burroughs Wellcome Fund, the Alfred P. Sloan Foundation, the Simons Foundation, the McKnight Foundation, the James S. McDonell Foundation, and the Office of Naval Research.}
\thanks{The authors are with the Departments of Electrical Engineering (A.D., B.A.), Applied Physics (S.G.), and Statistics (A.H.W.), Stanford University, Stanford, CA 94305 USA (e-mail: ahwillia@stanford.edu).}%
}
\date{}


\maketitle

\begin{abstract}
\noindent
Identifying recurring patterns in high-dimensional time series data is an important problem in many scientific domains.
A popular model to achieve this is convolutive nonnegative matrix factorization (CNMF), which extends classic nonnegative matrix factorization (NMF) to extract short-lived temporal motifs from a long time series.
Prior work has typically fit this model by multiplicative parameter updates---an approach widely considered to be suboptimal for NMF, especially in large-scale data applications.
Here, we describe how to extend two popular and computationally scalable NMF algorithms---Hierarchical Alternating Least Squares (HALS) and Alternatining Nonnegative Least Squares (ANLS)---for the CNMF model.
Both methods demonstrate performance advantages over multiplicative updates on large-scale synthetic and real world data.
\end{abstract}



\section{Introduction}

NMF models a matrix of nonnegative data, $\mX$, as the product two low rank and nonnegative matrices, thus approximating each datapoint (a row or column of $\mX$) as a conical combination of basis features or latent factors \autocite{Lee1999-nmf-learn-parts, Gillis2014-nmf-how-why}.
When the low rank assumption is appropriate, NMF often yields highly interpretable descriptions of data and thus is a highly effective tool for exploratory data analysis.
NMF has been applied to high-dimensional time series data, with applications ranging from audio processing, image processing, neuroscience, and text mining \autocite{Smaragdis2003-nmf-app-music, Jia2009, Gillis2014-nmf-how-why, Liu2017}.

However, many time series contain short-term temporal correlations or sequences of events that are not approximately low rank, and thus cannot be extracted by NMF.
For example, audio recordings are typically represented and visualized as spectrograms, which display the frequency content of sound over time as the signal varies.
Many sounds of interest have recognizable signatures in the frequency domain, which are not low rank---e.g. phonemes in human speech data may slightly change in frequency (pitch) over their production interval.
Similarly, in time series data from neuroscience, it is common to find clusters of brain cells that fire in a rapid sequence \autocite{Hahnloser2002, fujisawa2008behavior}.
NMF could efficiently model these firing events if neurons fired simultaneously; however, sparse sequences of neural firing yield high rank data matrices.
At best, NMF can only describe such sequences through multiple latent factors.
At worst, including additional factors to fit such structure may result in overfitting. 

\textit{Convolutive NMF} (CNMF) is a simple extension of the NMF model that overcomes these shortcomings.
As its name suggests, CNMF introduces convolutional structure into the low rank model reconstruction, and thus captures short-term temporal dependencies in time series data \autocite{Smaragdis2004-cnmf-deconv, Smaragdis2007-cnmf-speech-bases}.
The CNMF model has been effective in a variety of applications, including neuroscience \autocite{Mackevicius2018-cnmf-app-seqnmf}, medical data mining \autocite{Ramanarayanan2011-cnmf-app-mri}, and audio signal processing \autocite{Zhou2014-cnmf-app-phonemes}.

In recent years, algorithms for NMF have matured to a stage where it is computationally tractable to fit very large datasets \autocite{Kannan2016, Erichson2018}.
However, the CNMF model cannot be viewed as a special case of NMF, and thus these algorithmic improvements are not immediately transferable to CNMF.
As a result, while many high-performance and computationally scalable code packages are available for NMF \autocite{Zupan2012, scikit-learn}, algorithms and implementations of CNMF are less mature.

The Multiplicative Update (MU) algorithm appears to be the most common optimization routine for CNMF in published literature \autocite{Smaragdis2004-cnmf-deconv, Smaragdis2007-cnmf-speech-bases, Mackevicius2018-cnmf-app-seqnmf, Zhou2014-cnmf-app-phonemes}.
This method was originally developed for NMF \autocite{Lee1999-nmf-learn-parts, Lee2001-nmf-mult-alg}, and later adapted to CNMF \autocite{Smaragdis2004-cnmf-deconv, Smaragdis2007-cnmf-speech-bases}.
However, subsequent work found MU to be relatively ineffective for NMF \autocite{Gillis2014-nmf-how-why}, suggesting that MU may also be suboptimal for CNMF.
Here we derive two new algorithms for the CNMF model---Hierarchal Alternating Least Squares (HALS) and Alternating Nonnegative Least Squares (ANLS)---both of which can be understood as extensions of successful NMF algorithms \cite{Gillis2014-nmf-how-why} and are special cases of coordinate and block coordinate descent \autocite{Kim2014-nmf-unified-bcd, Wright2015}.
We show that HALS and ANLS outperform MU on CNMF models fit to large-scale data.
Additionally, we derive several reformulations of the CNMF objective function which lead to new and useful interpretations of the model.

\section{Background}

\begin{figure*}[t]
    \centering
  \subfloat[]{%
       \includegraphics[width=0.45\linewidth]{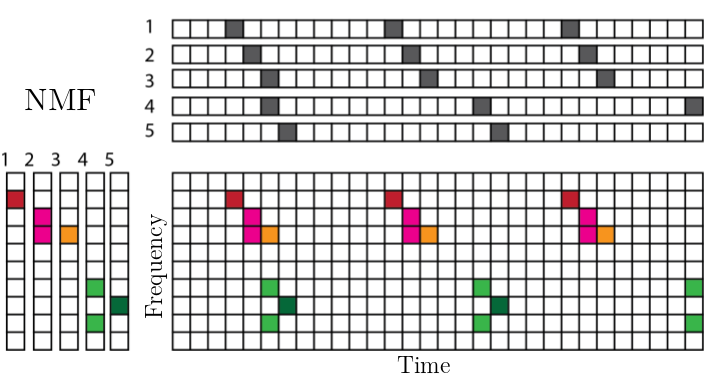}}
    \hfill
  \subfloat[]{%
        \includegraphics[width=0.45\linewidth]{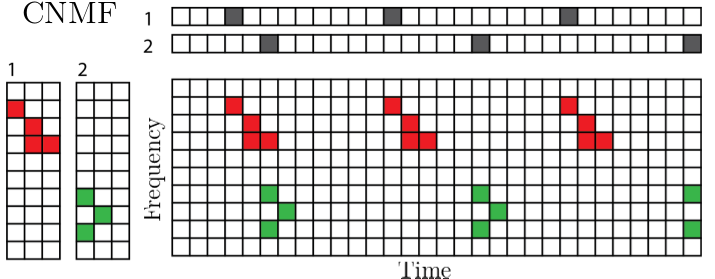}}
  \caption{
      Schematic illustration of NMF and CNMF models fitting the same dataset.
      (a) NMF models a data matrix $\mX$ (lower right) the product of $\mW$ (matrix with $K=5$ columns, left) and $\mH$ (matrix with $K=5$ rows, top). In the ``sum-of-outer-products'' interpretation of the model, each column of $\mW$ represents a group of simultaneously activated features, while the corresponding row of $\mH$ represents the times at which this group of features is active.
      (b) CNMF extends this ``sum-of-outer-products'' interpretation using a convolution operator instead of a vector outer product.
      Here, the same data are modeled using a tensor $\tW$ (tensor with $K=2$ slices, left) and $\mH$ (matrix with $K=2$ rows, top).
      Each of the $K$ slices of $\tW$ can be thought as a spatiotemporal feature of temporal duration $L$, and the times at which each such feature is convolutionally activated are specified by the corresponding row of $\mH$. This structure enables a much more compact and interpretable representation of this example time series.
    }
  \label{fig: nmf-cnmf-visual} 
\end{figure*}


\subsection{Notation}

We denote a vector with $P$ real-valued entries as $\vect{x} \in \bbR^P$, a $P \times Q$ matrix as $\mat{X} \in \bbR^{P \times Q}$, and a $P \times Q \times R$ tensor (in this paper, a tensor is an array with three indices) as $\tensor{X} \in \bbR^{P \times Q \times R}$. If a matrix (or vector or tensor) has strictly nonnegative entries, we write $\mat{X} \in \bbR^{P \times Q}_+$, or alternatively $\mat{X} \geq 0$.

We denote the $i$th slice of the tensor $\tensor{X}$ along its first mode as $\mat{X}_{i::}$, which indicates the first index is fixed to $i$ while the rest remain free.
In our example above, $\mX_{i::}$ refers to a matrix of size $Q \times R$, whereas $\mX_{::i}$ refers to a matrix of size $P \times Q$.
In the following sections, we overload the notation $\mat{W}_\ell = \mW_{\ell::}$ to refer to the matrix created by fixing the index of the tensor $\tensor{W}$ along the first mode. Concretely, if $\tensor{W}$ is a tensor of size $L \times N \times K$, then $\mat{W}_\ell$ is a matrix of size $N \times K$.

The symbol $\odot$ refers to element-wise multiplication of matrices, i.e. $(\mat{A} \odot \mat{B})_{ij} = \mat{A}_{ij} \mat{B}_{ij}$ (Hadamard product). Similarly, the notation $\frac{\mat A}{ \mat B}$ refers to element-wise division. In all cases, the \textit{norm} of a vector, matrix, or tensor is defined as the root sum-of-squares. For a tensor $\tensor{X} \in \bbR^{I \times J \times K}$, this is
\begin{align*}
    \| \tensor{X} \| = \left( \sum_{i=1}^I \sum_{j=1}^J \sum_{k=1}^K \tensor{X}_{ijk}^2 \right)^{1/2}.
\end{align*}

The symbol $\otimes$ refers to the Kronecker product between two matrices. If $\mat{A} \in \bbR^{m \times n}$ and $\mat{B} \in \bbR^{p \times k}$, then the Kronecker product $\mat A \otimes \mat B \in \bbR^{mp \times nk}$ is
\begin{align*}
    \mat A \otimes \mat B
    = \begin{bmatrix}
    \mat A_{11} \mat B & \hdots & \mat A_{1n} \mat B  \\
    \vdots & \ddots & \vdots \\
    \mat A_{m1} \mat B & \hdots & \mat A_{mn} \mat B
    \end{bmatrix}.
\end{align*}

If the matrix $\mat A$ has columns $\vect a_1, \hdots, \vect a_n$, then the vectorization of $\mat A$ is defined as
\begin{align*}
    \Vectorize(\mat A) = \begin{bmatrix} \vect a_1 \\ \vdots \\ \vect a_n \end{bmatrix}.
\end{align*}

When discussing update rules that solve the NMF and CNMF problem, we use superscripts to denote the iteration number of the algorithm. For example, $\mat W^{(i)} $ refers to the matrix $\mat W$ at the $i$th iteration of an algorithm or update scheme, $\mat W^{(i+1)}$ refers to the next iterate, and so on.


\subsection{The NMF and CNMF models}

Given a data matrix $\mX \in \bbR^{N \times T}_+$, NMF attempts to find two nonnegative factor matrices $\mW \in \bbR^{N \times K}_+$ and $\mH \in \bbR^{K \times T}_+$ that roughly approximate $\mX$. Formally, the NMF problem is:
\begin{equation} \label{eq: nmf objective}
\begin{aligned}
& \underset{\mW, \mH}{\text{minimize}}
& & \| \mX - \mW \mH \|^2 \\
& \text{subject to}
& & \mW \geq 0, \mH \geq 0.
\end{aligned}
\end{equation}
This model can be reformulated as a sum of outer products.
Letting $\vw_1, ..., \vw_K$ be the columns of $\mW$ and $\vh_1^\top, ..., \vh_K^\top$ be the rows of $\mH$, the objective is:
\begin{equation}
\begin{aligned}
& \underset{ \substack{\vw_1, \hdots, \vw_k \\\vh_1, \hdots, \vh_k}}{\text{minimize}}
& & \| \mX - \sum_{k=1}^K \vw_k \vh_k^\top \|^2 \\
& \text{subject to}
& & \vw_k \geq 0, \vh_k \geq 0 \quad \forall k \in \{1, 2, ..., K\}.
\end{aligned}
\end{equation}

The NMF model can be effective when $\mX$ is nonnegative and approximately low rank.
However, NMF may perform poorly as a feature extraction method when $\mX$ contains short-lived temporal motifs with high-rank structure.
This is demonstrated schematically in Figure \ref{fig: nmf-cnmf-visual}a.
The data matrix $\mat{X}$ represents a time series with $T$ measurements, with each column representing a single measurement of $N$ variables.
For example, $N$ could be the number of frequency bins in a spectrogram representation of an audio signal \cite{Smaragdis2007-cnmf-speech-bases}, or the number of recorded cells in a neural time series \cite{Mackevicius2018-cnmf-app-seqnmf}.
These time series can contain short-lived patterns that are not low-rank (in Fig. \ref{fig: nmf-cnmf-visual}, two different recurring patterns are shown in shades of red and green).
This results in NMF requiring many dimensions (i.e. a large choice for $K$) to capture the structure in the data.
This hampers interpretability as visible patterns in the data are split across multiple factors.

The convolutive NMF (CNMF) model was developed to address this shortcoming \autocite{Smaragdis2004-cnmf-deconv}.
CNMF finds a matrix $\mH \in \bbR_+^{K \times T}$ and a tensor $\tW \in \bbR_+^{L \times N \times K}$ that minimizes the following objective:
\begin{equation}
\begin{aligned}
& \underset{\tW, \mH}{\text{minimize}}
& & \|\mX - \sum_{\ell=1}^{L} \mW_{\ell} \mH \mS_{\ell-1} \|^2 \\
& \text{subject to}
& & \tW \geq 0, \mH \geq 0,
\end{aligned}
\label{eq: cnmf_objective}
\end{equation}
where $\mS_\ell$ is a $T \times T$ column right-shift matrix, defined as a matrix with ones along the $\ell$th upper diagonal and zeros otherwise. 
If $\vect{e}_i$ denotes the $i$th standard basis vector, then $\vect{e}_i^\top \mS_\ell = \vect{e}_{i+\ell}^\top$ when $i+\ell \leq T$.
A visual demonstration makes the role of $\mS_\ell$ clear:
\begin{align*}
\mat A = \begin{bmatrix}
1 & 2 & 3 & 4 \\
5 & 6 & 7 & 8
\end{bmatrix}
,&  \quad
\mat A \mS_1 = \begin{bmatrix}
0 & 1 & 2 & 3  \\
0 & 5 & 6 & 7
\end{bmatrix}
\\
\mat{A} \mS_2 = \begin{bmatrix}
0 & 0 & 1 & 2 \\
0 & 0 & 5 & 6
\end{bmatrix}
,& \quad \hdots
\end{align*}

When $L=1$, CNMF reduces exactly to NMF. 
Like ordinary NMF, CNMF also has a natural ``sum of outer products'' form. If we consider the slices $\mW_{::1}, ..., \mW_{::K} \in \bbR_+^{L \times N}$ and the row vectors $\vh_1^\top, \hdots, \vh_K^\top$ of the matrix $\mH$, we can define the \textit{convolution operator} $*$ by
\begin{align}
    \mat A = \mWk^\top * \vh_k^\top,
    \quad \mat A_{nt} = \sum_{\ell=1}^L \mW_{\ell n k} \mH_{k, t-\ell}.
\end{align}
which can alternatively be written as
\begin{align} \label{eq: matrix convolution}
    \mWk^\top * \vh_{k}^\top = \sum_{\tau = 1}^T \mH_{k\tau} \begin{bmatrix} \mat{0}_{\tau - 1} & \mWk^\top & \mat{0}_{T+1 - L - \tau} \end{bmatrix}.
\end{align}
where $\mat{0}_p$ signifies $p$ columns of zeros. 
To make the notation more concise, we abbreviate the zero-padding as follows:
\begin{align}
    \label{eq: pad notation}
    \left[ \mWk^\top \right]_\tau 
        &= \begin{bmatrix} \mat{0}_{\tau - 1} & \mWk^\top & \mat{0}_{T+1 - L - \tau} \end{bmatrix} \\
    \mWk^\top * \vh_{k}^\top 
        &= \sum_{\tau = 1}^T \mH_{k\tau}  \left[ \mWk^\top \right]_\tau
\end{align}
Using equation \ref{eq: matrix convolution}, we rewrite the CNMF objective as
\begin{equation}
\begin{aligned} \label{eq: cnmf-outer}
& \underset{\tW, \mH}{\text{minimize}}
& & \| \mX - \sum_{k=1}^{K} \mWk^\top * \vh_k^\top  \|^2 \\
& \text{subject to}
& & \tW \geq 0, \mH \geq 0.
\end{aligned}
\end{equation}
Here, each $\mW_{::k} \in \bbR^{N \times L}$ represents a short-lived temporal pattern, or \textit{motif}, that may have full rank.
The nonzero entries of each $\vh_k \in \bbR^{T}$ represent the times at which this motif occurs.
For the idealized time series in Figure \ref{fig: nmf-cnmf-visual}, CNMF pulls out a simpler and more interpretable description of data than NMF.
In essence, CNMF extracts 2 recurring patterns, corresponding to $K=2$ factors in the model.
In contrast, NMF requires $K=5$ model factors.

We note here briefly that different boundary conditions could be specified for the convolution operation in eq. \ref{eq: cnmf-outer}.
We adopted zero-padding for these boundary conditions as it appears to be the most standard choice in prior literature \autocite{Smaragdis2007-cnmf-speech-bases, Mackevicius2018-cnmf-app-seqnmf}.
Only minor modifications to our exposition would be needed to handle different choices.
For example, $\HH$ could be re-specified as a $K \times (T- L)$ matrix and each $\mS_\ell$ could be specified as a $(T - L) \times T$ matrix to specify convolution without padding.


\subsection{Multiplicative Update (MU) Algorithms}

The objective of the NMF problem (equation \ref{eq: nmf objective}) is non-convex, and finding an exact solution is NP-hard in general \autocite{Vavasis2010-nmf-complexity}. This had led to extensive algorithmic research on NMF, producing several effective heuristic algorithms \autocite{Gillis2014-nmf-how-why} and conditions guaranteeing an exact solution in polynomial time \autocite{Donoho2004-nmf-separable, Arora2016-nmf-provable}.

One such heuristic algorithm for NMF is the Multiplicative Update (MU) algorithm. The MU algorithm repeatedly updates $\mW$ and $\mH$ according to the following update rule \autocite{Lee2001-nmf-mult-alg}
\begin{align}
    \mW^{(i+1)}
    &= \mW^{(i)} \odot 
    \frac{\mX \mH^{(i)\top} }{ 
            \left[ \mW^{(i)} \mH^{(i)} \right] \mH^{(i) \top} },
\end{align}
where the index $i$ refers to the current iteration of the algorithm. By the symmetry of the NMF problem (the objective can be expressed as $\| \mX^T - \mH^T \mW^T \|$), the same update rule can be applied to $\mH$. In reality, the MU algorithm is actually just gradient descent with per-parameter scaling factors \autocite{Kim2014-nmf-unified-bcd}. Its popularity stems from several desirable properties---the update rule is monotonic, simple to implement, and preserves nonnegativity.


Since NMF is a special case of CNMF (with $L=1$), solving the latter problem exactly is also NP-hard. Accordingly, heuristic algorithms are also used to fit CNMF, most notably a generalization of MU \autocite{Smaragdis2007-cnmf-speech-bases}. In this case, the update rules are
\begin{align}
    \mWl^{(i+1)} = \mWl^{(i)} \odot \frac{\mX (\mH^{(i)} \mS_\ell)^\top}{\widehat \mX^{(i)} (\mH^{(i)} \mS_\ell)^\top }, \\
    \mH^{(i+1)} = \mH^{(i)} \odot \frac{ \sum_{\ell=1}^L \mWl^{(i) \top} \mX \mS_{-\ell}  }{ \sum_{\ell=1}^L \mWl^{(i) \top} \widehat\mX^{(i)} \mS_{-\ell} },
\end{align}
where $\widehat \mX^{(i)} = \sum_{\ell = 1}^L \mWl^{(i)} \mH^{(i)} \mS_{\ell -1}$ is our reconstruction of $\mX$. 
As in the NMF case, MU is easy to implement and has been applied frequently to fit CNMF. 
Nevertheless, past work has shown MU to be suboptimal for fitting NMF compared to other coordinate descent algorithms \autocite{Gillis2014-nmf-how-why, Kim2014-nmf-unified-bcd}. 
We reasoned that exploiting similar coordinate and block-coordinate updates would lead to performance benefits in the case of CNMF.

\subsection{Hierarchical Alternating Least Squares (HALS) for NMF}

Hierarchical alternating least squares (HALS) is a coordinate descent method used to fit NMF \autocite{Gillis2014-nmf-how-why, Cichocki2007-nmf-hals}. 
Each update step solves a constrained optimization problem exactly for a single column of $\mW$ or row of $\mH$. 
To update a single column $\vw_p$, we reformulate the NMF objective as 
\begin{align} 
    J(\mW, \mH) 
    &= \left \|  \mX - \sum_{k=1}^{K} \vw_k \vh_k^\top \right \|^2 
        \nonumber \\
    &= \left \|  \left( \mX - \sum_{k \neq p} \vw_k \vh_k^\top \right) - \vw_p \vh_p^\top  \right \|^2
        \label{eq: nmf-hals-reformulation}
\end{align}
and fix all variables except for the $p^\text{th}$ column of $\mW$.\footnote{As in the case of MU, the symmetry of the NMF problem allows us to use the same rule for $\mH$. For a more detailed derivation of the HALS updates for NMF, see \autocite{Cichocki2007-nmf-hals}.}
Minimizing over $\vw_p$ is a convex problem, and the Karush-Kuhn-Tucker (KKT) conditions for optimality generate the closed-form update rules \autocite{Cichocki2007-nmf-hals}:
\begin{align} \label{eq: nmf-hals-update-rule}
    \vw_p^{(i+1)} = \max \left( 0, \frac{ \left[ \mX - \sum_{k \neq p} \vw_k^{(i)} \vh_k^{(i) \top} \right] \vh_p^{(i)}} {\|\vh_p^{(i)}\|^2} \right ).
\end{align}
Numerical experiments suggest this update rule notably outperforms MU~\autocite{Gillis2014-nmf-how-why}. One possible explanation for this is that although both algorithms have a similar flop count to update all of $\mW$, HALS  solves many exact problems whereas MU computes a single, inexact gradient step.


\subsection{Alternating Nonnegative Least Squares (ANLS) for NMF}

Another popular approach to the NMF problem is to fix $\mat W$ or $\mat H$, and to solve the resulting convex sub-problem exactly.
This leads to an algorithm known as \textit{Alternating Nonnegative Least Squares (ANLS)}, whose updates are:
\begin{align} \label{ANLS-for-NMF}
    \mat W^{(i+1)} &= \argmin_{\mat W \geq 0} \|\mat X - \mat W \mat H^{(i)}\|^2, \\
    \mat H^{(i+1)} &= \argmin_{\mat H \geq 0} \|\mat X - \mat W^{(i + 1)} \mat H\|^2.
\end{align}
Each of these updates amounts to solving a \textit{nonnegative least squares} problem, which has been extensively studied in the optimization literature  \cite{Lin2007-nmf-proj-grad-anls, polyak2015projected-grad-nnls}.
Thus, one can leverage existing nonnegative least squares solvers to compute the solution to each sub-problem. 
A variety of such solvers are available, including active-set methods, quasi-newton methods, and projected gradient methods \autocite{Gillis2014-nmf-how-why}. 
This motivates us to also extend the ANLS approach to fit CNMF.


\section{New Algorithms for CNMF} 

\subsection{HALS} \label{sec: hals-for-cnmf}

In this section, we demonstrate how to extend HALS to fit the CNMF model, highlighting the key reformulations used in deriving the update rule.

\subsubsection*{Updating $\tW$}
Recall the CNMF objective from (\ref{eq: cnmf_objective}).
The sum $\sum_\ell \mWl \mH \mS_{\ell-1}$ can be written as a block matrix  product by defining
\begin{align*}
    \widetildemat \mW = \begin{bmatrix} \mW_1 & \mW_2 & ... & \mW_L \end{bmatrix},
    \quad \quad
    \widetildemat \mH = \begin{bmatrix} \mH\mS_0 \\ \mH\mS_1 \\ \vdots \\ \mH\mS_{L-1} \end{bmatrix}.
\end{align*}
Using the fact that $\sum_\ell \mWl \mH \mS_{\ell-1} = \widetildemat \mW \widetildemat \mH$, we can reformulate the CNMF objective as
\begin{equation}
\begin{aligned} \label{eq: cnmf-block-minimize}
& \underset{\tW, \mH}{\text{minimize}}
& & \| \mX - \widetildemat \mW \widetildemat \mH  \|^2 \\
& \text{subject to}
& & \widetildemat \mW \geq 0, \widetildemat \mH \geq 0, \\
& & & \widetildemat \mH_{\ell K : (\ell+1)K,:} = \widetildemat \mH_{0:K,:} \mS_\ell \\
& & & \text{ for all } \ell = \{0, \ldots, L-1 \},
\end{aligned}
\end{equation}
where the last constraint ensures that $\widetildemat \mH$ has the block matrix structure described above. This reveals an important fact: the CNMF approximation is an NMF factorization with linear constraints on $\widetildemat \mH$.

Due to this reformulation, it is clear that the HALS update rule for NMF extends easily to $\tW$.
When updating $\tW$, we treat $\mH$ as fixed, and thus we can ignore the linear constraints in \ref{eq: cnmf-block-minimize}.
Letting $\widetildevec \vw_p \in \bbR^N$ be the $p$th column of $\widetildemat \mW$ and $\widetildemat \vh_p \in \bbR^T$ be the $p$th row of $\widetildemat \mH$, we have the update rule
\begin{align}
    \widetildevec \vw_{p}^{(i+1)} 
    \coloneqq 
    \max \left( 
    0, \frac{ \left[ \mX  - 
            \sum_{j \neq p} \widetildevec \vw_{j}^{(i)} \widetildemat \vh_{j}^{(i) \top}  \right] \widetildemat \vh_{p}^{(i)} }{ 
            \| \widetildemat \vh_p^{(i)} \|^2 }
    \right).
\end{align}
This rule allows us to update $\vw_{\ell, :, k}$ using $p= (l-1)L + k$. 
Note that in practice, the matrices $\widetildemat \mH, \widetildemat \mW$ do not need to be explicitly instantiated. Each $\widetildevec \vw_p$ is simply an array view into $\tW$, and each block matrix $\HH \mS_\ell$ comprising $\widetildemat \HH$ can be computed on demand.

\subsubsection*{Updating $\mH$}
Deriving an update rule for $\mH$ is more complicated due to the convolutive structure imposed on $\mH$.
We first consider the outer product form of the CNMF objective from equation \ref{eq: cnmf-outer} and expand the convolution operator
\begin{align}
    &\quad \left \| \mX - \sum_{k=1}^{K} \mWk^\top * \vh_k^\top  \right \|^2 \nonumber \\
    &=
    \left \| \mX - \sum_{k=1}^K \sum_{\tau=1}^T \mH_{k\tau} \left[ \mWk^\top \right]_\tau \right \|^2 \nonumber \\
    \label{eq: cnmf-hals-full-expansion}
    &= \left \| \mat{E}^{(i)}
    - \mH_{kt} \left[ \mWk^\top \right]_t \right \|^2,
\end{align}
where $[\mW_{::k}^\top]_t$ is the matrix $\mW_{::k}^\top$ padded with $t-1$ columns of zeros on the left and $T+1-L-t$ columns of zeros on the right (first defined in (\ref{eq: pad notation}),
and where we define $\mat E^{(i)}$ as
\begin{align}
    \mat{E}^{(i)} 
        &= \left( \mX^{(i)} - \sum_{(p, \tau) \neq k, t}     \mH_{p\tau}^{(i)} \left[ \mW_{::p}^{(i)\top} \right]_\tau \right)_{:, t:t+L-1}.
\end{align}
This equation is reminiscent of equation \ref{eq: nmf-hals-reformulation}, and indeed leads to a related update rule.
Fixing all variables but a single entry $\mH_{kt}$, we can derive the Lagrangian and corresponding Karush-Kuhn-Tucker (KKT) conditions for optimality (see Appendix \ref{apdx: hals-cnmf-deriv}). 
This leads us to a closed form update rule for a single entry of $\mat H$:
\begin{align}
    \label{eq: cnmf-hals-updateh}
    \mH_{kt}^{(i+1)}
        &= \max \left( 0,  \frac{ \Tr( \mWk^{(i)}  \mat{E}^{(i)} ) }{ 
        \| \mWk^{(i)} \|^2 } \right),
\end{align}
which completes the generalization of HALS to CNMF.

Indeed, when $L = 1$, the HALS update rule for NMF (eq. \ref{eq: nmf-hals-update-rule}) can be recovered exactly.
Specifically, when $L=1$, both $\mat E^{(i)}$ and $\mat{W}_{::k}^{(i)}$ reduce to length-$N$ vectors: $\mat{E}^{(i)}$ is column $t$ of the residual matrix, and $\mat{W}_{::k}^{(i)}$ is the $k^\text{th}$ low rank factor.
Thus, the numerator term $\Tr( \mWk^{(i)}  \mat{E}^{(i)} )$ reduces to a vector inner product.
To update an entire row of $\mat H$ at once, as is standard for HALS in NMF, the numerator term may be extended to be a matrix-vector product, recovering eq. \ref{eq: nmf-hals-update-rule}.

However, when $L > 1$, updating the full row of $\mat H$ in closed form is not feasible.
Specifically, the update rule for $\mat H_{kt}$ is dependent on the current value of $\mat H_{k\tau}$ for all $t < \tau < t + L$, meaning that one can only simultaneously update every $L^\text{th}$ entry in the $k^\text{th}$ row in $\mat H$.
Thus, there are two potential extensions of HALS for CNMF, when updating $\mat H$:
\begin{itemize}
    \item \textit{Update $\mat H$ in blocks of size $T / L$}. Iterate over $\ell = 1, \hdots, L-1$ and, starting at position $\ell$, update every $L^\text{th}$ entry of row $k$ in $\mat H$. In principle, this could be achieved by appropriately truncating and reshaping the residual matrix $\mat E^{(i)}$.
    \item \textit{Update single entries of $\mat H$}. Iterate over $t = 1, \hdots, T$ and update $\mat H_{kt}$ by equation \ref{eq: cnmf-hals-updateh}. Note that one need not compute the full residual matrix; only columns ranging from $t$ to $t + L$ of $\mat E^{(i)}$ should be computed. This results in $O(NL)$ total floating point operations to update a single entry of $\mH$.
\end{itemize}
In both cases, the relevant entries in $\mat E^{(i)}$ should be updated after each parameter update.
The second option listed above (pure coordinate descent) is simpler to implement, and thus we focused on this variant in our numerical experiments \autocite{Kim2014-nmf-unified-bcd}.

As in HALS for NMF, adding $\ell_1$ regularization with weight $\alpha$ amounts to subtracting $\alpha$ from the numerator, and adding $\ell_2$ regularization with weight $\beta$ amounts to adding $\beta$ to the denominator \cite{Cichocki2007-nmf-hals}.
These extensions to the above algorithm could be used to identify regularized and sparse CNMF models.
As we are primarily interested in computational performance, we did not explore the statistical benefits of such regularization methods in detail. 

\subsection{ANLS}
In this section, we derive an Alternating Nonnegative Least Squares update rule for CNMF. We'll make use of two different formulations of the CNMF model, one for the update of $\tW$ and one for the update of $\mH$.

\subsubsection*{Updating $\tensor W$}
If we fix all entries of the matrix $\mat H$, updating the tensor $\tensor W$ using ANLS is straightforward. We recall the formulation from (\ref{eq: cnmf-block-minimize}), in which we expressed the CNMF model as a product of block matrices: $\widehat{\mat X} = \widetildemat{\mW} \widetildemat{\mH}$.
In this form, it is clear that we can update the matrix $\widetildemat{\mW}$ using an off-the-shelf Nonnegative Least Squares solver. Since the matrix $\widetildemat{\mW}$ is simply a reshaped version of the tensor $\tW$, this suffices for updating $\tW$. Concretely, we have the following update rule:
\begin{align}
    \widetildemat{\mW}^{(i+1)} = \argmin_{\widetildemat{\mW} \geq 0} \|\mX - \widetildemat{\mW} \widetildemat{\mH}^{(i)}\|^2.
\end{align}


\subsubsection*{Updating $\mH$}
The update of $\mH$ requires us to use a different formulation of the CNMF model. First, we recall the following fact (see, e.g., \autocite{Horn-matrix-analysis}):
\begin{align} \label{eq: kronecker-identity-1}
    \Vectorize(\mat A \mat B \mat C)
    = (\mat C^\top \otimes \mat A) \Vectorize(\mat B),
\end{align} 
for any matrices $\mat A, \mat B, \mat C$ (assuming appropriate dimensions).
This leads us to the following vectorized version of the CNMF model:
\begin{align*}
    \Vectorize(\widehat{\mX}) 
    &= \underbrace{\sum_{\ell=1}^L (\mS_{\ell-1}^\top \otimes \mWl)}_{\mat V} \Vectorize(\mH). 
\end{align*}
When $\tW$ is fixed, the above equation allows us to update $\mH$ by solving a single Nonnegative Least Squares problem, as we did in the case of $\tW$. 
With this definition, the ANLS update rule for $\mH$ is given by
\begin{align}
    \mH^{(i+1)} = \argmin_{\mH \geq 0} \| \Vectorize(\mX) - \mat V \Vectorize(\mH) \|^2.
    \label{eq: anls-H-update}
\end{align}
Thus, we have cast the CNMF optimization problem as an Alternating Nonnegative Least Squares problem. 
In practice, the matrix $\mat V \in \bbR_{+}^{NT \times KT}$ may be too large to fit in memory. 
One way around this is to use a matrix-free method, which requires access to the matrix $\mat V$ only through it's matrix vector product. 
For example, Projected Gradient Descent and Fast Iterative Shrinkage Thresholding (FISTA) are good candidate methods if efficient implementations of $\mat V \vect z$ and $\mat V^T \vect z$ are available \cite{Lin2007-nmf-proj-grad-anls, polyak2015projected-grad-nnls}. In practice, we find that directly solving (\ref{eq: anls-H-update}) at each iteration is inefficient. However, the formulation above leads to two insights.
    
Noting that $\mat V$ is a block-toeplitz matrix, it becomes clear that the update for $\mH$ is actually a higher-dimensional analogue to the standard nonnegative deconvolution problem studied in the literature \autocite{vogelstein2010fast}. The difference is that here, the coefficient matrix is block-toeplitz rather than toeplitz. This suggests the possibility of leveraging the convolutional structure of the problem using approaches which have been applied in the deconvolution case. We leave this to future work.

The second insight is that updating a single column of $\mH$ with the other columns held fixed is simply a Nonnegative Least Squares problem in $K$ variables which does not require explicitly storing the matrix $\mat V$. Therefore, one approach to solving (\ref{eq: anls-H-update}) is block coordinate descent, updating a single column at a time. Since block coordinate descent converges to the optimal solution for Nonnegative Least Squares problems \autocite{beck2013convergence-bcd}, this approach would eventually reach the optimal solution for (\ref{eq: anls-H-update}). In practice, it is not necessary to exactly solve (\ref{eq: anls-H-update}) at each iteration. For the purposes of our numerical experiments, we make a single pass of coordinate descent at each iteration (updating each column exactly once), using the block-principal pivoting method described in \autocite{kim2011fast-anls}.


\section{Numerical Experiments}

\begin{figure*}[t]
    \centering
    \includegraphics{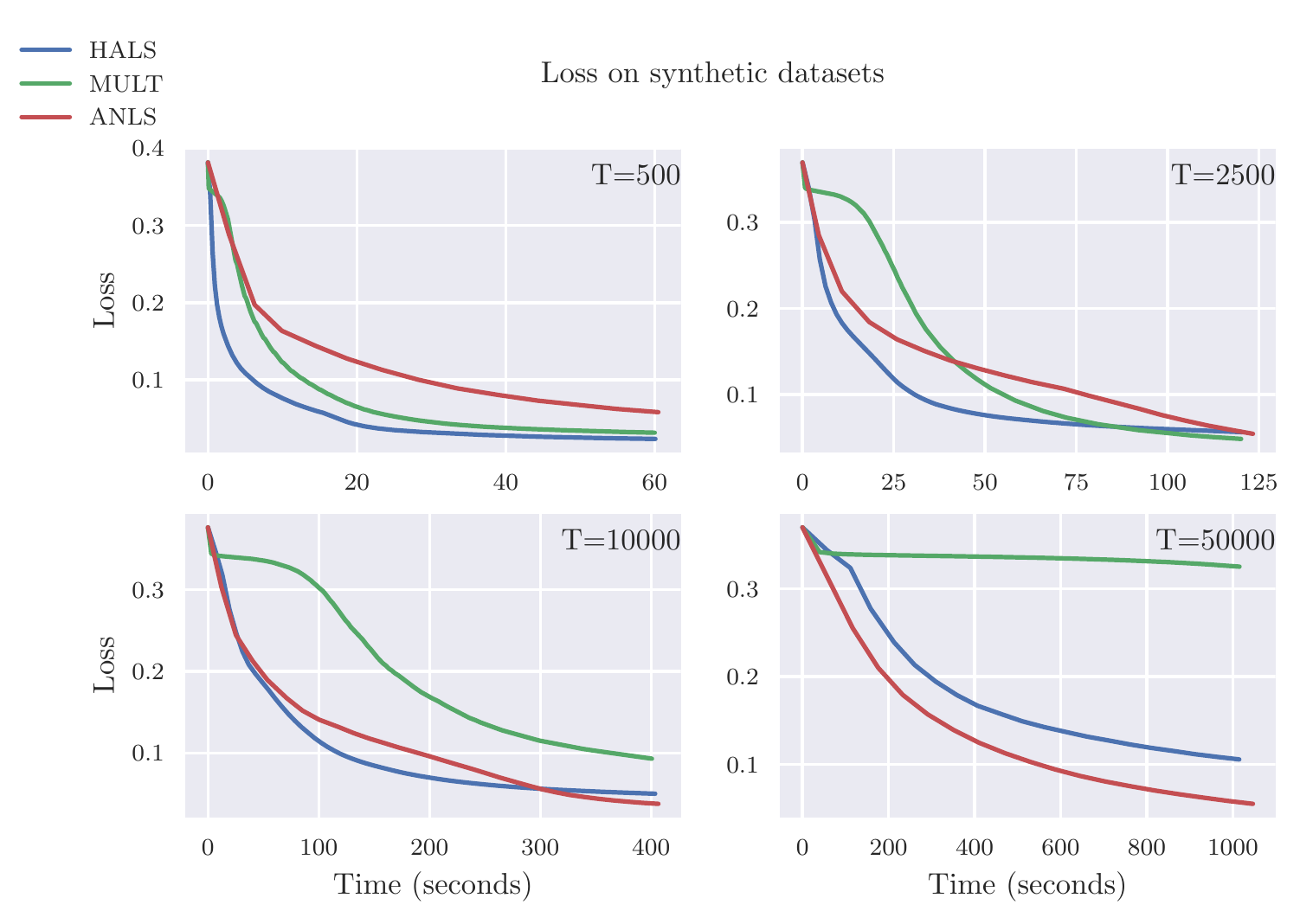}
    \caption{
    Algorithm performance on synthetic data. 
    The vertical axis denotes normalized loss of the CNMF model, $\| \mX - \widehat{\mX} \|   / \| \mX \|$; the horizontal axis denotes cumulative computation time.
    As dataset size increases (denoted by number of timebins $T$) the performance of HALS and ANLS improves relative to multiplicative updates.
    For $T=500$ and $T=2500$ all three algorithms perform similarly.
    For $T=10000$, multiplicative updates takes significantly longer to converge.
    Finally, for $T=50000$, multiplicative updates makes little to no progress in the allotted time (1000 seconds), whereas both ANLS and HALS rapidly converge.
    }
    \label{fig: synthetic-comparison}
\end{figure*}

\begin{figure}[t]
    \centering
    \includegraphics[width=\columnwidth]{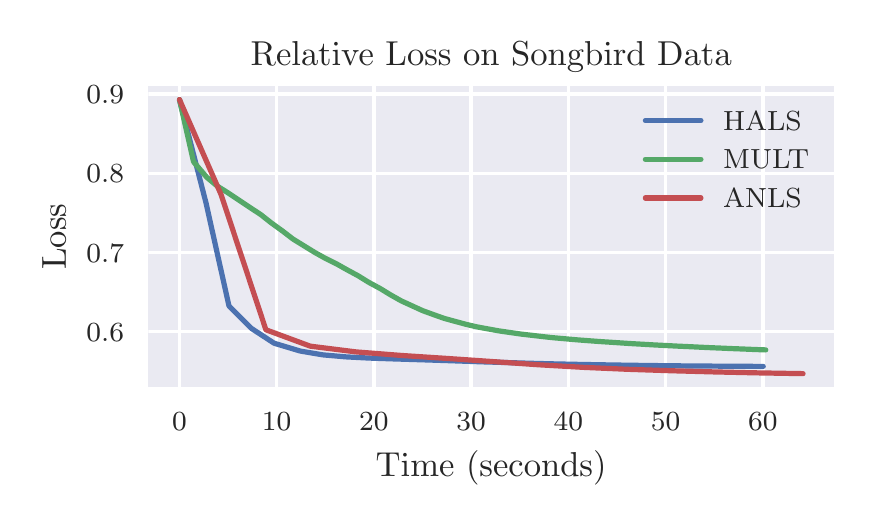}
    \caption{
        Algorithm performance on a songbird spectrogram from \autocite{Mackevicius2018-cnmf-app-seqnmf}. 
        ANLS and HALS perform similarly and nearly converge after 20 seconds; multiplicative updates takes approximately three times as long to achieve the same objective value.
    }
    \label{fig: songbird}
\end{figure}

\begin{figure*}[t]
    \centering
    \includegraphics{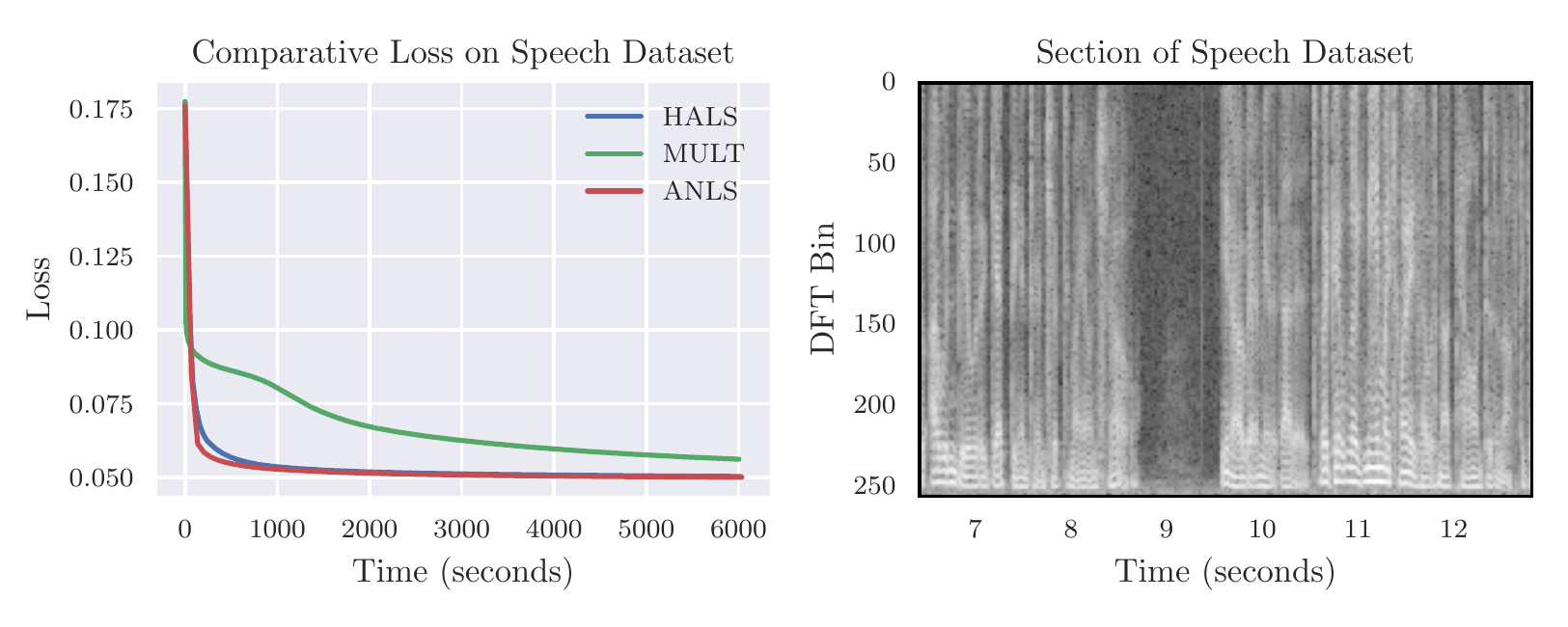}
    \caption{
    Comparison of algorithms on a large speech dataset. \textbf{Left:} Normalized loss, $\| \mX - \widehat{\mX} \|   / \| \mX \|$, achieved by each algorithm as a function of computation time.
    Both HALS and ANLS converge significantly faster than multiplicative updates.
    \textbf{Right:} a small slice of the speech dataset, representative of the full recording.
    }
  \label{fig: speech-comparison}
\end{figure*}

\begin{figure*}[t]
    \centering
    \includegraphics[width=\linewidth]{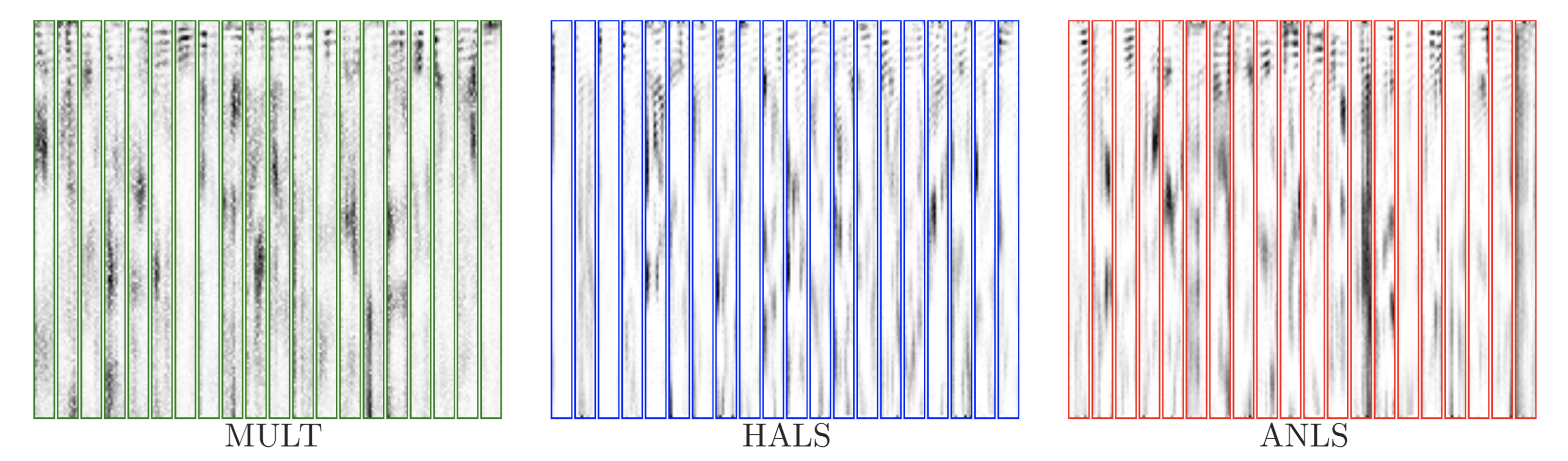}
    \caption{
        The twenty components $W_{::k}$ recovered by multiplicative updates, HALS, and ANLS.
        The vertical axis spans frequency (DFT bin) and the horizontal axis spans time.
        One rectangle surrounded by a colored border is a single $W_{::k}$.
        The borders of each component are colored to match the previous loss plots.
        For all three algorithms, the components are perceptually similar but appear in different orders.
        Specifically, each algorithm recovers harmonic stacks that correspond to different sounds frequently spoken during the recording.
    }
    \label{fig:speech_bases}
\end{figure*}

In this section, we compare all three algorithms on synthetic and experimental data.
We find that HALS and ANLS both converge significantly faster than MU, and that their relative performance to MU increases with dataset size. 
For example, on a large audio dataset, we find that HALS converges roughly five times faster than MU. 
This effect occurs consistently regardless of random initialization.

In each figure, we measure reconstruction error (loss) by the scaled norm of the residual, $\| \mX - \widehat{\mX} \|   / \| \mX \|$. 
All results are obtained via the Julia \autocite{julia} code using version 1.0, published in the GitHub repository at \url{github.com/degleris1/CMF.jl}, which contains implementations of all algorithms and Jupyter notebooks to reproduce figures. 
We use the Sherlock compute cluster at Stanford to run all simulations, using two cores (Broadwell) with 16GB of memory per core. In all experiments, all algorithms were given the same random initialization.

\subsection{Synthetic Data}

In this experiment, we test each algorithm on synthetic data of various sizes.
The synthetic datasets were generated from a CNMF model with added noise, as follows:

\begin{itemize}
    \item The dimensional parameters were chosen to be $N = 250$, $L = 20$, $K = 5$. We generated and examined four otherwise identical datasets with $T = 500$, $T = 2500$, $T = 10000$, and $T = 50000$.
    \item Each $\vw_{:nk}$ (the length-$L$ fibers of $\tensor{W}$) followed a randomly shifted Gaussian curve. Specifically, let $f(\tau; \mu_{nk}, \sigma)$ denote a univariate Gaussian probability distribution function with mean $\mu$ and standard deviation $\sigma$. We set $\sigma = 0.2$ and sampled $\mu_{nk}$ uniformly at random between $-1$ and $1$.
    We then randomly sampled amplitude parameters from a symmetric Dirichlet distribution $\boldsymbol{\alpha}_n \sim Dir(0.1)$, achieving approximately sparse vectors $\boldsymbol{\alpha}_n \in \mathbb{R}^K_+$ representing loadings across each component.
    Finally, we set $\tensor{W}_{\ell nk} = \alpha_{nk} f(2\ell/L - 1; \mu_{nk}, \sigma)$, for $\ell = 1, \hdots, L$.
    This procedure was repeated for each feature $n = 1, \hdots, N$ and component $k = 1, \hdots, K$.
    \item Each element in $\mH$ was set to zero with probability 0.1, and otherwise randomly sampled from an exponential distribution with a rate parameter $\lambda=1$. Similar to our construction of $\tensor{W}$, this produced a synthetic dataset with sparse factors, in agreement with previously reported results on real data (e.g. \cite{Mackevicius2018-cnmf-app-seqnmf}).
    \item 
    The \textit{ground truth} matrix is given by $\mX^{\text{true}} = \sum_{\ell=1}^L \mWl \mH \mS_{\ell-1}$. 
    We then added truncated Gaussian noise, $\mX_{nt} = \max(0, \mX^{\text{true}}_{nt} + e_{nt})$ where each $e_{nt}$ was drawn uniformly from a standard normal distribution (zero-mean and unit standard deviation). 
    The matrix $\mX$ was given as input to all algorithms.
\end{itemize}

Convergence on synthetic data is shown in Figure \ref{fig: synthetic-comparison}. For small dataset sizes, all three algorithms give similar performance. As dataset size grows, however, HALS and ANLS converge much more quickly than MU. This is best illustrated when $T = 50000$ columns. On this large dataset, MU fails to converge within the 1000 second limit.

\subsection{Results on a Songbird Spectrogram}

In this experiment, we fit the CNMF model on a songbird spectrogram from \autocite{Mackevicius2018-cnmf-app-seqnmf} (available at \url{github.com/FeeLab/seqNMF}). The dimensions of the data matrix are $141$ DFT bins (rows) by $4440$ timebins (columns), and we use a motif length of $L=50$ and $K=3$ factors.
The timebins are sampled at 200~Hz.
We run each algorithm for 60 seconds and plot the relative loss over time.
We find that both HALS and ANLS converge after around 20 seconds, whereas multiplicative updates fails to converge within the 60 second time limit (Fig. \ref{fig: songbird}).
All algorithms find perceptually similar components (data not shown).

\subsection{Qualitative Results on a Large Speech Dataset}
\label{section:speech}

In this experiment, we fit the CNMF model on a large dataset consisting of two males speaking as part of an interview. 
Following the procedure in \cite{OGrady-2007discovering}, we down-sample the audio recording to 8KHz and compute a magnitude spectrum using an FFT window of 512 samples, and an overlap of 384.
This yields a data matrix of size $257 \times 20149$ which we fit using $K=20$ components and motif length of $L=12$ time-steps. 
As a final preprocessing step, we log-transform the spectrogram and add a constant (so that all entries are nonnnegative).

We observe that HALS and ANLS converge to their final loss roughly 5x faster than MU.
A small section of the magnitude spectrogram, along with a convergence comparison, is shown in Figure \ref{fig: speech-comparison}.

A natural question is whether the components found by HALS and ANLS are similar to those found by MU. 
We find that this is indeed the case. Figure \ref{fig:speech_bases} shows that components recovered by all three algorithms are perceptually similar, each containing distinctive horizontal bands which correspond to the harmonics found in human speech. 
The components extracted in this experiment look similar to those found by \cite{OGrady-2007discovering}.

\section{Conclusion}
In this paper, we have shown how to extend two popular algorithms for NMF, HALS and ANLS, to the Convolutive NMF problem.
Both algorithms offer faster convergence rates than MU, with speedups of around 5x noted on a large dataset, and were observed to recover qualitatively similar motifs.
In situations where the practitioner must perform a parameter search over regularization strengths or the number of motifs, this speedup is of practical value.
Future research could investigate improvements to the ANLS algorithm by incorporating specialized nonnegative least squares solvers and potentially exploiting the block Toeplitz structure of eq. \ref{eq: anls-H-update}.
To handle even larger datasets, randomized variants of the CNMF algorithms described here could also be developed, in analogy to recently proposed randomized variants of HALS in NMF \autocite{Erichson2018}.
Overall, we expect these improvements to enable convolutional factor modeling on a variety of high-dimensional time series data with much longer durations than what has been previously explored.


\appendices
\section{Reformulations of the CNMF objective}

The \textit{classical form} of the CNMF approximation is
\begin{align} \label{eq: apdx-classical-form-cnmf}
    \mat f (\tW, \mH)
    &= \sum_{\ell=1}^L \mWl \mH \mS_{\ell-1}
\end{align}

We define the \textit{convolution operator} as
\begin{align}
    \mWk^\top * \vh_k^\top
    = \sum_{\tau=1}^T \mH_{k\tau} [\mWk^\top]_\tau
\end{align}
where $[\mWk^\top]_\tau = \begin{bmatrix} \mat{0}_{\tau - 1} & \mWk^\top & \mat{0}_{T+1-L-\tau} \end{bmatrix}$ and $\mat{0}_p$ is a $N \times p$ matrix of zeros.
This allows us to write the \textit{outer product form} of the CNMF approximation
\begin{align}
    \mat f(\tW, \mH)
    &= \sum_{k=1}^K \mWk^\top * \vh_k^\top
\end{align}

Another useful formulation comes from considering Kronecker identities. Given three matrices $\mat A, \mat X, \mat B$, we know
\begin{align} \label{eq: kronecker-identity}
    \Vectorize(\mat A \mat X \mat B)
    = (\mat B^\top \otimes \mat A) \Vectorize(\mat X)
\end{align}
from \autocite{Horn-matrix-analysis}. This leads us to the \textit{Kronecker form} of the CNMF approximation, which is
\begin{align}
    \Vectorize(\mat f(\tW, \mH))
    &= \sum_{\ell=1}^L (\mS_{\ell-1}^\top \otimes \mWl) \Vectorize(\mH) \\
    &= \sum_{\ell=1}^L (\mS_{1-\ell} \otimes \mWl) \Vectorize(\mH)
\end{align}
We define the matrix $\mat V = \sum_{\ell=1}^L \mS_{1-\ell} \otimes \mWl$, which is also written as
\begin{align}
    \mat V
    &= \begin{bmatrix}
    \mW_1   & \mat 0    & \hdots    & \mat 0 \\
    \mW_2   & \mW_1     & \hdots    & \mat 0 \\
    \vdots  & \vdots    &           & \vdots \\
    \mW_L   & \mW_{L-1} & \hdots    & \mat 0 \\
    \mat 0  & \mW_{L}   & \hdots    & \mat 0 \\
    \vdots  & \vdots    &           & \vdots \\
    \mat 0  & \mat 0    &           & \mW_{1}
    \end{bmatrix}
\end{align}
where $\mat 0$ is a $T \times T$ matrix of zeros. Thus the Kronecker form is concisely written as $\Vectorize(\mat f(\tW, \mH)) = \mat V \Vectorize(\mH)$.

Alternatively, we can take the transpose of equation (\ref{eq: apdx-classical-form-cnmf}) and apply (\ref{eq: kronecker-identity}) to write the \textit{Toeplitz form} of the CNMF approximation
\begin{align}
    \Vectorize(\mat f(\tW, \mH)^\top )
    &= \sum_{l=1}^L (\mWl \otimes \mS_{1-\ell}) \Vectorize(\mH^\top)
\end{align}
which is also written as
\begin{align}
    \Vectorize(\mat f(\tW, \mH)^\top )
    &= \begin{bmatrix}
    \mathcal{T}(\vw_{:11}) & \hdots & \mathcal{T}(\vw_{:1K}) \\
    \vdots & \ddots & \vdots \\
    \mathcal{T}(\vw_{:N1}) & \hdots & \mathcal{T}(\vw_{:NK})
    \end{bmatrix}
\end{align}
where $\mathcal T(\vect v) \in \bbR^{T \times T}$ is a Toeplitz matrix defined for any vector $\mathbf{v} \in \bbR^L$ as
\begin{align}
    \mathcal T(\vect v)
    &= \begin{bmatrix}
    \vect v_1   & 0             & \hdots    & 0\\
    \vect v_2   & \vect v_1     &           & 0 \\
    \vdots      & \vdots        &           & \vdots \\
    \vect v_L   & \vect v_{L-1} & \hdots    & 0 \\
    0           & \vect v_L     & \hdots    & 0 \\
    \vdots      & \vdots        &           & \vdots \\
    0           & 0             & \hdots    & \vect v_1
    \end{bmatrix}
\end{align}
i.e. the $\ell$th diagonal below the main diagonal is equal to $\vect v_{\ell+1}$.

\section{Derivations of the HALS update rules}

\subsection{HALS for NMF}

Consider the NMF objective, written as
\begin{equation*}
\begin{aligned}
& \underset{ \substack{\vw_1, \hdots, \vw_k \\\vh_1, \hdots, \vh_k}}{\text{minimize}}
& & \left\| \mX - \sum_{k=1}^K \vw_k \vh_k^\top \right\|^2 \\
& \text{subject to}
& & \vw_k \geq 0, \vh_k \geq 0 \quad \forall k \in \{1, 2, ..., K\}
\end{aligned}
\end{equation*}
We will derive a closed-form update rule that updates a single column $\vw_k$ or a single row $\vh_k^\top$.
By the symmetry of the problem, it suffices to derive this update rule for $\vw_k$ only.
First choose $k$ and let $\mat E = \mX - \sum_{p \neq k} \vw_p \vh_p^\top$.
Our minimization problem is now
\begin{equation*}
\begin{aligned}
& \underset{\vw_k}{\text{minimize}}
& &  \left\| \mat E - \vw_k \vh_k^\top \right\|^2 \\
& \text{subject to}
& & \vw_k \geq 0
\end{aligned}
\end{equation*}
Applying the identity $\| \mat X \|^2 = \Tr (\mX^\top \mX)$, we can rewrite $J(\vw_k) = \left\| \mat E - \vw_k \vh_k^\top \right\|^2$ as
\begin{align*}
    J(\vw_k)
    &= \Tr \left( \mat E^\top \mat E \right) + 
        \Tr( \vh_k \vw_k^\top  \vw_k \vh_k^\top ) -
        2 \Tr( \mat E^\top \vw_k \vh_k^\top ) \\
    &= \Tr \left( \mat E^\top \mat E \right) + 
        \Tr( \vh_k^\top \vh_k \vw_k^\top  \vw_k  ) - 
        2 \Tr( \vh_k^\top \mat E^\top \vw_k  ) \\
    &= \| \mat E \|^2 + 
        \| \vh_k \|^2 \| \vw_k \|^2 -
        2 \vh_k^\top \mat E^\top \vw_k 
\end{align*}
Next, we write the Langrangian as
\begin{align*}
    \mathcal L (\vw_k, \vect \lambda)
    = \Tr \left( \mat E^\top \mat E \right)
        + \| \vh_k \|^2 \| \vw_k \|^2
        - 2 \vh_k^\top \mat E^\top \vw_k
        - \vect \lambda^\top \vw_k
\end{align*}
which has gradient
\begin{align*}
    \nabla_{\vw_k} \mathcal L (\vw_k, \vect \lambda)
    =  2 \| \vh_k \|^2 \vw_k
        - 2 \mat E \vh_k
        - \vect \lambda
\end{align*}
Setting this equal to zero gives us the KKT conditions
\begin{align}
    \vw_k
        &= \frac{\mat E \vh_k + \frac{1}{2} \vect \lambda}{\|\vh_k\|^2} \\
    \vect \lambda &\geq 0 \\
    \vw_k &\geq 0 \\
    \label{eq: nmf-complement-slackness}
    (\vw_k)_i \vect \lambda_i &= 0 \quad \forall i = 1, 2, \hdots, N
\end{align}
If $(\mat E \vh_k)_i \geq 0$, then we must have $\vect \lambda_i = 0$ to satisfy equation (\ref{eq: nmf-complement-slackness}). 
If $(\mat E \vh_k)_i \geq 0$, then we must have $(\vw_k)_i = 0$. 
This leads to the closed form solution
\begin{align}
    \vw_k
        &= \max\left( 0, 
                \frac{\mat E \vh_k}{\|\vh_k\|^2} 
            \right)
\end{align}

\subsection{HALS for CNMF} \label{apdx: hals-cnmf-deriv}

For the CNMF model, the HALS update rule loses its symmetry across $\tW$ and $\mH$.
However, as demonstrated in Section \ref{sec: hals-for-cnmf}, the update rule for $\tW$ can be derived using the HALS update rule for NMF.
To derive the update rule for $\mH$, we begin by choosing $k, t$ and defining $\mat E = \mX - \sum_{(p, \tau) \neq (k, t)} \mH_{p, \tau} \left[ \mW_{::p}^\top \right]_\tau$. From (\ref{eq: cnmf-hals-full-expansion}), we can update $\mH_{kt}$ with the optimization problem
\begin{equation}
\begin{aligned}
\label{eq: optimize-H}
& \underset{\mH_{kt} }{\text{minimize}}
& & J(\mH_{kt}) 
        = \left\| \mat E - \mH_{kt} [ \mW_{::k}^\top]_t \right\|^2 \\
& \text{subject to}
& & \mH_{kt} \geq 0
\end{aligned}
\end{equation}
Since $[\mW_{::k}^\top]_t$ only interacts with $L$ columns of $\mat E$, we can define $\mat R = \mat E_{:, t+L-1}$ and write (\ref{eq: optimize-H}) as
\begin{equation}
\begin{aligned}
& \underset{\mH_{kt} }{\text{minimize}}
& & J(\mH_{kt}) 
        = \left\| \mat R - \mH_{kt} \mW_{::k}^\top \right\|^2 \\
& \text{subject to}
& & \mH_{kt} \geq 0
\end{aligned}
\end{equation}
Since $\mH_{kt}$ is just a scalar, it is quite simple to derive a closed form update rule.
The corresponding Lagrangian is
\begin{align}
    \mathcal L (\mH_{kt}, \lambda)
        &= \| \mat R \|^2 
            + \mH_{kt}^2 \| \mW_{::k}^\top \|^2 \\
        &\quad - 2 \mH_{kt} \Tr ( \mW_{::k} \mat R )
            - \lambda \mH_{kt} \nonumber
\end{align}
which has gradient
\begin{align}
    \nabla_{\mH_{kt}} \mathcal L (\mH_{kt}, \lambda)
        &=  2 \mH_{kt} \| \mW_{::k}^\top \|^2 \\
        &\quad - 2  \Tr ( \mW_{::k} \mat R )
            - \lambda \nonumber
\end{align}
This gives us the KKT conditions
\begin{align}
    \mH_{kt} &= \frac{ \Tr ( \mW_{::k} \mat R ) + \frac{1}{2} \lambda }{ \| \mW_{::k}^\top \|^2 } \\
    \mH_{kt} &\geq 0 \\
    \lambda &\geq 0 \\
    \lambda \mH_{kt} &= 0 \label{eq: HALS complementary slackness}
\end{align}
Equation (\ref{eq: HALS complementary slackness}), which is referred to as the complementary slackness condition, implies that either $\lambda$ or $\mH_{kt}$, must be zero. This allows us to update $\mH_{kt}$ using the closed form update rule from Section \ref{sec: hals-for-cnmf}.




\printbibliography

\end{document}